\documentclass[11pt]{article}

\usepackage[preprint]{acl}

\usepackage{times}
\usepackage{latexsym}
\usepackage{todonotes}

\usepackage[T1]{fontenc}
\usepackage[utf8]{inputenc}
\usepackage{microtype}
\usepackage{inconsolata}

\usepackage{graphicx}
\usepackage{stfloats}          
\usepackage{afterpage}         
\usepackage{placeins}          

\title{ALMeD: Medieval French Gold-Standard}

\author{Ragini Menon and Sabine Tittel\\[3pt]
Heidelberg Academy of Sciences and Humanities\\
\texttt{ragini.menon@hadw-bw.de}, \texttt{sabine.tittel@hadw-bw.de}
}

\begin{document}
\maketitle

\begin{abstract}
Medieval French is one of the Romance vernacular languages with a significant history of transmission and an enormous influence on other vernacular languages of the time. The sources convey valuable knowledge about all aspects of culture. The study of Medieval French has, over the last few years, increasingly benefited from NLP-based methods.
There have been numerous efforts to parse medieval French; however, these efforts use corpora comprising legal texts and—primarily—literary texts as their foundation and training data. This means that, except for juridical resources, \textbf{scientific literature} is 
overlooked. 
We present \textbf{ALMeD}\footnote{A first version is available at \url{https://huggingface.co/datasets/HaDW-ALMA/ALMeD}.}, an annotated, semantically rich corpus (work-in-progress) of medical and surgical treatises produced in medieval French, i.e., Old and Middle French, including \textbf{Anglo-Norman}. Our \textbf{semantically disambiguated gold-standard} goes beyond existing resources, focusing on \textbf{medical terminology}, and accounts for morphological development, evolving syntactical structures and lack of spelling normalisation in these languages.
\end{abstract}

\section{Introduction}
\label{sec:intro}

Natural Language Processing (NLP) for historical language stages remains a niche area of research, due to the lack of resources and challenges caused by non-standard syntax and spelling, among others. Nevertheless, recent years have seen a growing number of initiatives, such as Universal Dependencies and various efforts within the Digital Humanities for various languages like Ancient Greek, Latin, Classical Chinese, and Ancient Hebrew. However, there are still major issues, such as the scarcity of high-quality training data, especially gold-standard annotated corpora; difficulties in contextualising information accurately; and issues such as model hallucination.

One such language that suffers from these issues is Medieval French — a broad categorisation that includes Old French and Middle French. Old French in itself refers to a group of language varieties spoken primarily in Northern France from around the 9th Century until the middle of the 14th Century, and followed by the Middle French language period, which lasted until circa 1500.

The rise of textual culture and new educational institutions in Medieval cities led to the emergence of new expert groups and methods of knowledge generation and transmission, fostering the use of vernacular languages, including Old French (and other Old Gallo-Romance languages such as Old Occitan, as well as Old Italo-Romance language varieties) as a form of written communication. While (Middle) Latin remained dominant in administration, education, and scholarship for a long time, vernaculars began to play an increasingly important role — albeit in regionally varied ways — often still dependent on Latin models.

This linguistic shift was gradual and uneven, reflecting deeper socio-cultural transformations. The adoption of vernaculars for conveying knowledge changed intellectual networks and signalled a broader renewal of the intellectual landscape, contributing to the development of Europe as a knowledge society. The conceptual development of these languages both reflected and drove social change, forming a crucial part of Europe’s cultural and scientific heritage.

Within the Romance vernaculars, Medieval French was a dominant medium of cultural life, encompassing not only literature but also art and architecture, technical developments and medicine, amongst other things. The prestige of French in the medieval world is directly reflected in the number of text witnesses from Northern France, England, the Holy Land, etc. The textual tradition comprises many thousands of texts.
This makes developing NLP-supported approaches to analysing Medieval French texts — for philological, linguistic, and cultural-historical research questions — crucial.

However, although some NLP tools already exist for Old French — such as the \textit{Stanford CoreNLP} model, the \textit{Pyrrha} project \cite{clerice_deucalion_2018} by the École Nationale des Chartes, and the \textit{Syntactic Reference Corpus of Medieval French} (SRCMF) — significant gaps remain. In particular, there is a lack of high-quality training data enriched with detailed semantic annotation. While early experiments with training LLMs on Old French have produced encouraging results, they have been limited in both scale and depth.

\section{Related Work}
\label{sec:related work}

Studies on NLP for Medieval French, such as \citealp{grobol_is_2021} and \citealp{guibon_parsing_nodate}, have highlighted the difficulty of parsing Medieval French. Results from other studies, such as \citealp{kezerian_ye_2024}, also suggest that distinguishing between Medieval and Modern French is complicated.

The primary resource for experiments with Medieval French text has been the \textit{Syntactic Reference Corpus of Medieval French} (SRCMF) by \citealp{stein_syntactic_2013}. Originally a treebank, the SRCMF has been adapted to conform to the Universal Dependencies (UD) standards. The SRCMF dataset, however, is not comprehensive as it is primarily restricted to Part-of-Speech (POS) tagging and does not include annotations for lemmata. Studies such as \citealp{stein_parsing_2014} to compare dependency parsers and \citealp{grobol_bertrade_2022} to evaluate the usage of contextual word embedding models for downstream tasks such as POS tagging and dependency parsing have used the dataset as training data and achieved promising results, despite the limited features of the dataset.

For the Pyrrha\footnote{\url{https://dh.chartes.psl.eu/pyrrha/}} ecosystem by \citealp{camps_corpus_2021}, a tagger for Old French, the \textit{OFr3C} corpus was assembled from juridical texts from the \textit{Corpus juris civilis} (École Nationale des Chartes), a codicifation of Roman law, translated into French, and literary text genres such as the \textit{Chansons de Geste} and the \textit{Roman Courtois}; for the latter, the preexisting \textit{Chrétien} corpus, compiled for the \textit{Dictionnaire Électronique de Chrétien de Troyes} \cite{kunstmann_dictionnaire_2016}, was reused.

The part-of-speech annotations are complex, accounting for the various pronoun categories (personal, demonstrative, indefinite, interrogative, relative, adverbial, impersonal, cardinal, ordinal), and times, and mode and person are also specified for verbs; for a reference lexeme inventory, Pyrrha uses the \textit{Altfranzösisches Wörterbuch} by Tobler and Lommatzsch - TL \cite{TL}. They also included annotations for contractions, agglutinations and non-bounded terms, named entities and homonyms, among other details.
     \section{ALMeD}
\label{sec:dataset}
While existing corpora such as the \textit{OFr3C} corpus and the \textit{Base de Français Médiéval} - BFM, ENS de Lyon, Laboratoire IHRIM \cite{guillot_base_2018}, are available, they are relatively small and mainly focus on literary texts. In contrast, ALMeD is domain-specific. While its technical texts, of course, also contain non-technical language at large, they account for specialised terminology seldom covered by the \textit{belles lettres}.\footnote{With notable exceptions, e.g., we find sophisticated shipbuilding terminology in a hagiographical Anglo-Norman text from ca 1193 \cite{EdmK}, technical terms of several domains in the \textit{Roman de la Rose} from ca 1230/ca 1275 \cite{Doerr:Rose2}, etc.}
ALMeD focuses on medieval medical texts that contain highly complex medical and surgical terminology, which makes it suitable not only for NLP-related tasks of the \textit{Knowledge Networks in Medieval Romance Speaking Europe} (ALMA) project.\footnote{Based at the Heidelberg Academy of Sciences and Humanities, the Bavarian Academy of Sciences and Humanities, and the Academy of Sciences and Literature Mainz, \url{https://www.hadw-bw.de/alma} [2026-01-29].} project, but it also holds potential for research on technical texts more broadly. Although medical and surgical terminology is highly specialised, these texts share with other technical domains the communicative practices typical of technical language — practices that were shaped historically and transmitted as part of a broader tradition of discourse.
ALMeD focuses on medieval medical texts containing highly complex medical and anatomical terminology. While it is primarily designed to support NLP-related tasks within the ALMA project, it may also be of interest for research on technical texts more broadly. Although medical and surgical terminology is highly specialised, these texts share with other technical domains a range of communicative practices typical of technical language, such as standardised modes of description, instruction, and argumentation. These practices were historically established and transmitted over time, reflecting a broader tradition of technical discourse.
The gold-standard fine-grained annotated dataset will consist of texts from the medical corpus of the ALMA project. Using the \textit{Dictionnaire étymologique de l'ancien français} (DEAF, \citealp{DEAF}) as the primary lexicographic reference, these texts have been semantically disambiguated and annotated for several key linguistic features: part-of-speech, including the universal part-of-speech (UPOS) 
and the language-specific part-of-speech (XPOS), lemma, main lemma (standard headword form of the word), named entities, glossary definitions (of medical and surgical terms provided in the critical text editions), and dictionary definitions in modern French and English (see below).

\subsection*{Texts Used}

At the time of writing, the following texts are being annotated and validated:

\begin{enumerate}
    \item The \textit{Chappitre singulier} (an overview over the history of medicine and the medical society at the time) and the second treatise (on tumors, cancer, and the plague) of the \textit{Grande Chirurgie} by Gui de Chauliac, Middle French, (text after 1363) manuscript 2\textsuperscript{nd} third 15\textsuperscript{th} c., edition in preparation by ALMA; around 12,000 tokens - annotation in progress.
    \item The first treatise (on anatomy) of the same text, edition by Sabine Tittel \cite{tittel_anathomie_2004}; around 32,000 tokens - annotation completed, currently being corrected.
    \item \textit{The Surgery of Albucasis}, edition by David Trotter \cite{trotter_albucasis_2005}, Lorraine scripta, mid 13\textsuperscript{th} c.; around 130,000 tokens - annotation completed, currently being corrected.
    \item \textit{Lettre d'Hippocrate à César}, edition by Tony Hunt \cite{hunt_popular_1990}, Anglo-Norman scripta, 2\textsuperscript{nd} quarter 13\textsuperscript{th} c.; around 24,000 tokens - completed.
\end{enumerate}


\subsection*{Part of Speech Tagging}

For the part-of-speech (POS) tagging, two tagsets will be used. The first is Universal Part-of-Speech (UPOS) tags as per the Universal Dependencies (UD), which is a framework for consistent annotation of grammar across different languages. Using the UD format is particularly useful in standardising annotations for a project like ALMA that works with four languages. The second set of POS tags used are the language-specific tags (XPOS), which are even more fine-grained than the UPOS.

We chose the Universal Dependencies for two reasons:

\begin{enumerate}
    \item Since it is the accepted standard framework for linguistic annotation, we aimed to maximise the reusability of our data, particularly for downstream learning tasks, such as low-resource settings and multilingual training.
    \item In ALMA, we work with multiple languages, including Old Occitan and Old Italian. We also align parallel texts in two languages; for example, ChirAlbT has textual transmissions in Old French and Old Occitan, while GuiChaul has transmissions in Old Italian and Middle French. While XPOS is language-specific and fine-grained, we treat UPOS as a point of commonality. This is particularly relevant for our lexical-semantic studies, as it allows us to examine how a given word is tagged across languages: does it belong to a different part of speech in each language, and if so, why? Do other factors, such as historical context, account for these differences?
\end{enumerate}

For the language-specific morpho-syntactic (XPOS) tags, we use the CATTEX tagset \cite{guillot_base_2018}. The CATTEX tags comprise two fields: category and type. The categories mostly correspond to the classic parts of speech: VER (verb), NOM (noun), ADJ (adjective), PRO (pronoun), DET (determiner), ADV (adverb), PRE (preposition), CON (conjunction), INJ (interjection). Each linguistic unit has a single morphological label, composed of the category and type (except for the categories PRE, ETR, ABR, RED and OUT, which include only the category name). Labelling is morphosyntactic in the sense that categories and types are determined in context according to primarily morphological principles (strictly syntactic annotation is also carried out). We also account for named entities with two additional custom tags PROPNper (Proper Noun Person) and PROPNpla (Proper Noun Place).

\subsection*{Lemmatisation}

One of the key issues in NLP for Medieval French is lemmatisation, i.e., assigning variant graphical realisations of a lexeme to a form defined as a standard, called a lemma. The lemmatisation is a crucial step because of the lack of spelling normalisation in that period: A given lemma can be expressed in many different spellings and, on the other hand, a given spelling can be an expression of different lemmata (for challenges and implications, cf. \citealp[192-199]{Tittel:2024}). The lemmatisation thus also determines the POS of the lexeme
(and it also maps inflected verb occurrences to the appropriate lemma).

While there has been some success in developing a fairly accurate lemmatiser for medieval French, Pyrrha (which is based on the TL lemmata), the training data used by the developers is primarily focused on literary and juridical texts. This means that scientific texts and terminology have not been accounted for. To annotate the lemma, we use the lemma list of the DEAF as our primary reference, which contains around 84,000 lemmata.
This makes it the most comprehensive dictionary of Old French. The lemmatisation of the DEAF corresponds to a 12th-century French accepted as 'standard form' in the discipline; it therewith adopts the rules of the TL, but corrects them in line with modern practice, e.g. with regard to umlaut placement. However, in cases where the word is not listed in the DEAF, other dictionaries are referenced, such as the \textit{Anglo-Norman Dictionary} (AND)\footnote{\url{https://anglo-norman.net/}} and the \textit{Dictionnaire du Moyen Français} (DMF)\footnote{\url{http://zeus.atilf.fr/dmf/}}.

Since the DEAF is a dictionary that groups its entries by word families to make etymological relations visible, there is also the added benefit of including the main lemma — the lemma of the word family — in our lemmatisation process.

\subsection*{Glossary Definitions}

As this dataset focuses on medical terminology, we include glossary definitions of medical and anatomical terms drawn from the critical editions of the medieval manuscripts. The glossary is essential for sense definitions and, thus, for semantic disambiguation in context. For instance, 
in the GuiChaul\textsc{m}T edition \cite[299]{tittel_anathomie_2004}, there is a glossary entry, \textit{chief} (masculine), recorded with two senses depending on context:
\begin{itemize}
\item 1\textsuperscript{o} "upper part of the human body, head", attested in lines 319; 333; 339; 346; 350; 356;
358; 359; 360; 370; etc. 
\item 2\textsuperscript{o} "terminal part of something (of a bone, etc.); tip", attested in lines  201; 219; 320
\end{itemize}

The correct contextual usage of this word in both definitions is indicated in the Glossary by the references to the lines in which the word occurs, which makes it a vital resource for semantic disambiguation. There are many such instances in which such general and domain-specific terms have several meanings; therefore, these instances will be tagged with the appropriate definition.
Since there is currently no all-encompassing resource with a complete lexeme inventory and state-of-the-art genus-differentia definitions of all senses that could serve as a centralised dataset, this approach helps fill in lexicographic gaps. This is particularly important in all cases where the corresponding dictionary entry in the DEAF is part of DEAF\textit{pré}.\footnote{For the differences of DEAF\textit{pré} and DEAF\textit{plus}, see \citealp[48]{Tittel_2023_ceci}.}

\subsection*{Dictionary Definitions}

We also include the dictionary definitions of key words and terms from the various Old French to Modern French and Old French to Modern English dictionaries, to resolve semantic ambiguity as much as possible. 

\subsubsection*{Medieval French to Modern French} For the Modern French sense definitions, we rely on five dictionaries: DEAF, AND, DMF, \textit{Tobler-Lommatzsch} (TL), and the \textit{Französisches Etymologisches Wörterbuch} (FEW). 
Among these, the DEAF serves as the primary lexicographic reference and provides the majority of our definitions. This preference reflects both ALMA’s conceptual continuity with the DEAF project and our privileged access to key resources such as its lemma list. AND, DMF and the other dictionaries are used in cases where the word is not listed in the DEAF.

\subsubsection*{Medieval French to Modern English} For the English sense definitions, we use two dictionaries: \textit{Anglo-Norman Dictionary}\footnote{\url{https://anglo-norman.net/}} and the \textit{Old French-English Dictionary} by \citealp{hindley_old_2006} for standard definitions. While neither dictionary is comprehensive nor do they contain detailed lexicographic and etymological information, these are the best medieval French-to-English resources available, and can be used for basic tasks.

\subsection*{Annotation Process}

\begin{figure*}[t] 
  \centering
  \includegraphics[width=\textwidth]{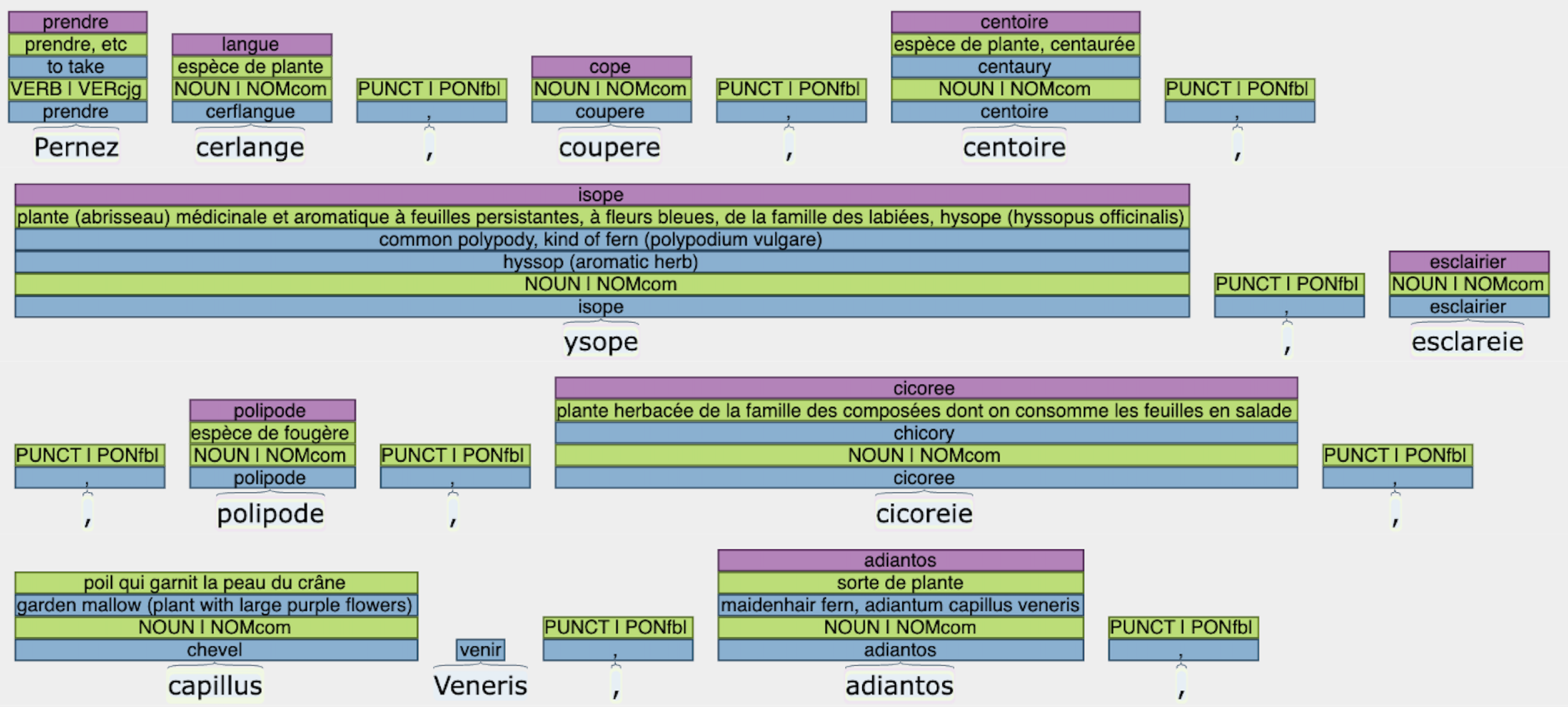}
  \caption{Annotating \textit{Lettre d’Hippocrate}, an Anglo-Norman text on INCEpTION}
  \label{fig:inception}
\end{figure*}

To annotate the texts, we use the semantic annotation platform INCEpTION \cite{klie_inception_2018}. While we initially considered other tools, such as TXM and CATMA, INCEpTION proved to be the most suitable for our objectives. Its capabilities—specifically designed for the creation of training data—include preset layers for Universal Dependencies, support for custom annotation layers (e.g. glossary and dictionary definitions), recommender systems that accelerate the annotation process, broad file-format compatibility (in particular XML/TEI), and built-in automatic tokenisation and sentence segmentation.

We have four standard layers for every token: part-of-speech (UPOS and XPOS), lemma and main lemma — a minimum of four labels per token. The glossary and dictionary definitions are tagged once; however, in situations where one word has two meanings, the token is assigned its contextually accurate definition (see Figure~\ref{fig:inception}). INCEpTION's layer functionalities also allow us to differentiate between the dictionary sources used, thereby making it easier to trace the definitions back to their origin, ensure consistency across annotations, and resolve cases where multiple resources provide overlapping or conflicting information.

\section{Future Work}

We aim to create a complete gold standard comprising upwards of 50,000 tokens annotated and texts semantically disambiguated, and present ALMeD as a suitable, domain-specific alternative to the SRCMF dataset. In contrast to existing resources, ALMeD will capture not only morphosyntactic information but also semantic and domain-specific layers, making it a richer dataset for the study of medieval medical French texts.

ALMeD will also be used internally by the ALMA project for, among other things, our NLP research, supporting the development and evaluation of new tools, models, and methodologies. ALMeD is also the foundation for a dedicated language model for Medieval French that we are currently developing, HeidelBERT.

The first version of ALMeD is publicly available on Hugging Face. As the annotation is still ongoing, we will progressively make more data available to the research community as a reusable and extensible resource for more accurate linguistic analysis, improved language models, and a solid foundation for future research, not only in medieval French linguistics but also in NLP for under-resourced languages.

\FloatBarrier        
\bibliography{custom}

\end{document}